\RequirePackage{iftex}\ifPDFTeX\pdfoutput=1\fi  
\documentclass[runningheads]{llncs}

\usepackage{eccv}
\usepackage{eccvabbrv}
\usepackage{graphicx}
\usepackage{booktabs}
\usepackage{array}
\usepackage{amsmath}
\usepackage{amssymb}
\usepackage{xcolor}
\usepackage{tikz}
\usetikzlibrary{arrows.meta,positioning,calc,fit}
\usepackage{hyperref}
\hypersetup{hidelinks}

\newcommand{\hota}{3D~HOTA}
\newcommand{\final}{62.3947}

\begin{document}
\setlength{\emergencystretch}{1em}

\title{Detection Is Not the Finish Line: Multiscale Decoding and
Constraint-Aware Global Association for Multi-Camera 3D Tracking}
\titlerunning{Multiscale Decoding and Constraint-Aware Global Association}

\author{Abdullah Naeem\inst{1,2} \and
Anav Katwal\inst{1,+} \and
Ayon Dey\inst{1,+} \and
Md Tamjidul Hoque\inst{1,*}}
\authorrunning{A. Naeem et al.}
\institute{LSU New Orleans\\
\email{\{anaeem,akatwal,aday,thoque\}@lsuneworleans.edu} \and
ParkLabs.ai\\[0.5ex]
\raisebox{0.5ex}{\scriptsize *}\hspace{0.15em}Corresponding author.\quad
\raisebox{0.5ex}{\scriptsize +}\hspace{0.15em}Equal contribution.}

\maketitle

\begin{abstract}
A query-based multi-view 3D detector can transfer well to a new calibrated
camera network and still produce badly fragmented tracks: on the AI City
Challenge 2026 Track~1 validation set, the frozen Sparse4D detector behind the
official challenge winner splits each Person identity into a median of eight
fragments, and half of the gaps between fragments exceed the horizon of a
causal tracker.  We show that most of the remaining headroom of such a system
lies after detection.  Keeping the public detector frozen, we add
(i)~class-wise source selection between two input scales, which recovers small
and distant people from a $0.75\times$ pass while leaving large objects at the
scale the detector was trained for; (ii)~a geometry-to-appearance bridge that
projects every 3D box into all calibrated views, matches it to 2D person
detections, and aggregates robust track-level re-identification descriptors;
and (iii)~constraint-aware consolidation, which combines a causal short-gap
stitch with offline centroid-linkage clustering under two hard constraints,
temporal exclusion and metric 3D reachability, plus class-aware long-gap
stitching for rigid objects.  Every association stage changes object
identities only.  On the challenge evaluation server the system scores
\final{} \hota{} (DetA 57.4501, AssA 56.8131, LocA 80.1272), first on the
general leaderboard snapshot of 30~October~2026 (second at the time of
writing, after a competitor's detector update) and 5.63 points above our
reproduction of the official winner; the constrained association stage alone
contributes +3.73 HOTA through a +5.49 AssA gain at unchanged DetA.
Controlled fixed-box validation studies, a constraint ablation, sensitivity
curves, and a catalogue of negative results support the design.
\end{abstract}

\section{Introduction}
\label{sec:intro}

Multi-camera 3D perception asks for a scene-level account of every object: a
position in a shared world frame and one identity for as long as the object
is present, however many cameras it crosses.  Track~1 of the 2026 AI City
Challenge~\cite{tang2026aicity,wang2026physicalai} poses this problem in
warehouses with up to 20 calibrated cameras, seven object classes, synthetic
and real scenes, and no depth at inference time.  The metric is a 3D
adaptation of HOTA~\cite{luiten2021hota}, whose factors separate detection
quality (DetA), association quality (AssA), and localization (LocA).

\begin{figure}[t]
  \centering
  \includegraphics[width=\textwidth]{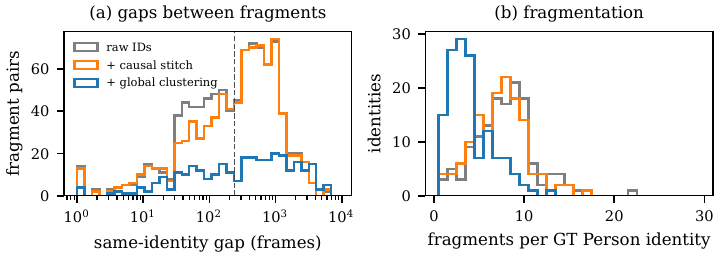}
  \caption{Why a causal tracker cannot repair the frozen detector's tracks.
  Validation Person predictions (half-resolution production boxes) are matched
  to ground truth per frame with a 1~m centre gate; a fragment is a predicted
  identity matched to one ground-truth person on at least ten frames.
  (a)~Gaps between consecutive fragments of the same person: with raw
  Sparse4D identities (grey) the median gap is 274 frames and 52\% of gaps
  exceed the 240-frame causal horizon (dashed); the causal stitch (orange)
  removes mostly short gaps, leaving 57\% of the remainder beyond the horizon.
  Constrained global clustering (blue) removes two thirds of all gaps.
  (b)~Fragments per ground-truth identity: median 8 (interquartile range
  6--10) raw, 7 after the causal stitch, 3 after global clustering.}
  \label{fig:gaps}
\end{figure}

The official winner of the track, GaugeAlign~\cite{wu2026gaugealign}, runs a
frozen NVIDIA TAO Sparse4D checkpoint~\cite{lin2023sparse4dv3} after a
closed-form canonicalization of the camera network's coordinate gauge, and won
on detection alone: its authors report a 10-point DetA margin over the
runner-up and a 7-point AssA deficit.  Our reproduction of that system scores
56.7690 \hota{} on the evaluation server with 55.7189 DetA but only 49.6755
AssA.  Figure~\ref{fig:gaps} quantifies the reason.  Matching the detector's
raw tracks to validation ground truth, each Person identity is covered by a
median of eight predicted identities, the median gap between consecutive
fragments of one person is 274 frames, and 52\% of those gaps are longer than
the 240-frame horizon of a causal stitcher; 18\% exceed 900 frames (30~s).
The tracklets themselves are about 95\% pure.  The detector, in other words,
is not the bottleneck; the topology of its tracks is.

This observation motivates a detector-preserving strategy.  Rather than train
another 3D detector, we ask two questions.  First, can a different input scale
recover the small, distant people the detector misses, without harming the
classes it already handles well?  Second, can fragmented world-space tracks be
consolidated over arbitrarily long gaps using appearance, while the physical
constraints available in a calibrated 3D scene prevent the over-merges that
make unconstrained re-identification (ReID) clustering unsafe?  Both answers
are affirmative, and together they move the server score to \final{} \hota{},
first on the general leaderboard snapshot of 30~October~2026
(Table~\ref{tab:leaderboard}).

Our contributions are:
\begin{itemize}
  \item \textbf{Class-wise multiscale source selection} for a frozen Sparse4D
  model (Sec.~\ref{sec:multiscale}).  Person boxes come from a $0.75\times$
  pass and every other class from the $0.50\times$ training scale; the output
  of each class is \emph{replaced}, never unioned, so no duplicates arise.
  Full-validation Person HOTA after association rises from 35.115 to 40.108,
  and the server score by 0.9243.
  \item \textbf{A geometry-to-appearance bridge} (Sec.~\ref{sec:bridge}) from
  world-space 3D boxes to multi-view Person descriptors: projection,
  one-to-one matching, confidence- and overlap-weighted cross-view fusion, and
  trimmed track-level prototypes.  It supplies descriptors for 98.6\% of the
  Person rows in the final test submission.
  \item \textbf{Constraint-aware consolidation} (Secs.~\ref{sec:short}--\ref{sec:rigid}):
  a causal short-gap stitch, an offline centroid-linkage clustering of whole
  tracks under temporal-exclusion and 3D-reachability constraints, and
  class-aware long-gap stitching for six rigid classes.  On the server this
  stage adds 3.7321 HOTA with DetA unchanged to four decimals.  An ablation
  shows that temporal exclusion, which refuses hundreds of above-threshold
  pairs of look-alike people seen simultaneously, is what makes a fixed
  moderate threshold safe, while the reachability gate is a cost-free
  safety margin.
  \item \textbf{Evidence.}  A seven-step server progression, fixed-box
  validation studies in which only identities change, a constraint ablation,
  sensitivity curves over the clustering hyperparameters, and a catalogue of
  nineteen negative results, all with an evaluator verified to reproduce the
  official TrackEval numbers exactly.
\end{itemize}

\section{Related Work}
\label{sec:related}

\paragraph{Multi-view 3D detection.}
Early multi-view detectors project per-camera features onto a ground plane
and reason in bird's-eye view~\cite{hou2020mvdet,teepe2024earlybird}; the
automotive literature replaced dense BEV grids with learned queries that
sample image features through calibration~\cite{wang2021detr3d,li2022bevformer}.
Sparse4D~\cite{lin2023sparse4dv2,lin2023sparse4dv3} keeps a recurrent bank of
sparse instance queries whose temporal propagation yields detection and
tracking in one model.  GaugeAlign~\cite{wu2026gaugealign} showed that the
learned world-coordinate anchors of NVIDIA's warehouse checkpoint of this
family~\cite{wang2026unified} are not translation equivariant across sites,
and repaired the failure with a label-free canonicalization of the
calibration.  We keep the frozen checkpoint and that correction, and address
the two error sources GaugeAlign left open: scale-dependent misses and
fragmented identities.

\paragraph{Query-based and online tracking.}
MOTR~\cite{zeng2022motr} and MUTR3D~\cite{zhang2022mutr3d} carry identity
implicitly in persistent queries; MOTIP~\cite{gao2025motip} instead predicts
identity labels from a trajectory memory.  SparseQID~\cite{shrestha2026sparseqid}
applies the latter to Sparse4D queries for this challenge and ranked third in
the official online track.  The SORT family~\cite{bewley2016sort,wojke2017deepsort,zhang2022bytetrack,aharon2022botsort,cao2023ocsort,du2023strongsort}
associates detections frame by frame with motion and appearance.  All of
these are causal; the stage that produces most of our gain clusters whole
tracks with future observations, so we do not claim the challenge's online
bonus.

\paragraph{Multi-camera association with appearance.}
Hierarchical clustering of tracklets under ReID affinity and spatio-temporal
constraints is the classical multi-camera design~\cite{zhang2017hierarchical},
refined at city scale by CityTrack~\cite{lu2023citytrack} and, in the 2025
edition of this challenge, by an offline global-tracking stage that the
winning team found decisive~\cite{tang2025aicity,ziovision2025aic}.  Our
setting differs in that the primitives are already fused 3D tracks from a
query-based detector, not single-camera tracklets.  We therefore project the
3D hypotheses back into every camera to collect appearance, and we impose
reachability in metric world coordinates rather than in image space.

\paragraph{Track~1 systems.}
Syn2RealTrack~\cite{tran2026syn2realtrack} (second in the official track)
lifts 2D detections through calibration and adapts to the synthetic-to-real
gap online; our own earlier system~\cite{naeem2026geometry} followed a similar
geometry-first route with estimated depth.  The present work abandons that
route in favour of the frozen 3D detector and concentrates on what happens
after it.

\section{Method}
\label{sec:method}

\begin{figure}[t]
  \centering
  \begin{tikzpicture}[font=\scriptsize, node distance=2.5mm and 2.5mm,
    box/.style={draw, rounded corners=1.5pt, align=center, inner sep=2pt,
                minimum height=7.5mm, fill=white, line width=0.5pt},
    det/.style={box, fill=blue!7, draw=blue!50!black},
    ass/.style={box, fill=orange!10, draw=orange!60!black},
    arr/.style={-{Latex[length=1.5mm]}, line width=0.5pt}]
    \node[box, text width=15mm] (rgb) {calibrated RGB,\\$N$ cameras};
    \node[det, right=of rgb, text width=20.5mm] (gauge) {gauge\\canonicalization\\(Eq.~\ref{eq:gauge})};
    \node[det, text width=24mm] (s50) at ($(gauge.east)+(15.5mm,4.6mm)$) {frozen Sparse4D\\$0.50\times$ ($540{\times}960$)};
    \node[det, text width=24mm] (s75) at ($(gauge.east)+(15.5mm,-4.6mm)$) {frozen Sparse4D\\$0.75\times$ ($810{\times}1440$)};
    \node[det, text width=30mm] (sel) at ($(gauge.east)+(45.5mm,0)$) {class-wise source selection\\Person $\leftarrow 0.75\times$ ($\tau{=}0.65$)\\all other classes $\leftarrow 0.50\times$};
    \node[ass, text width=23mm, minimum height=9.5mm] (bridge) at ($(rgb.west)+(11.5mm,-18.5mm)$) {3D$\to$2D bridge, robust\\ReID prototypes\\(Sec.~\ref{sec:bridge})};
    \node[ass, right=of bridge, text width=21.5mm, minimum height=9.5mm] (causal) {causal short-gap\\stitch, $\le 240$ frames\\(Sec.~\ref{sec:short})};
    \node[ass, right=of causal, text width=23mm, minimum height=9.5mm] (global) {constrained global\\clustering, no gap limit\\(Sec.~\ref{sec:global})};
    \node[ass, right=of global, text width=21.5mm, minimum height=9.5mm] (rigid) {rigid-class long-gap\\stitch\\(Sec.~\ref{sec:rigid})};
    \node[box, right=of rigid, text width=13mm, minimum height=9.5mm] (out) {scene-level\\3D tracks};
    \node[box, dashed, text width=22mm, below=2mm of bridge] (reid) {Person ReID encoder\\(trained once, train split)};
    \draw[arr] (rgb) -- (gauge);
    \draw[arr] (gauge.east) -- (s50.west);
    \draw[arr] (gauge.east) -- (s75.west);
    \draw[arr] (s50.east) -- (sel.west);
    \draw[arr] (s75.east) -- (sel.west);
    \draw[arr] (sel.south) -- ++(0,-2.2mm) -| (bridge.north);
    \draw[arr] (reid) -- (bridge);
    \draw[arr] (bridge) -- (causal);
    \draw[arr] (causal) -- (global);
    \draw[arr] (global) -- (rigid);
    \draw[arr] (rigid) -- (out);
    \node[font=\scriptsize, blue!50!black, anchor=south west] at ($(rgb.west |- s50.north)+(0,0.8mm)$) {detection: boxes change};
    \node[font=\scriptsize, orange!60!black, anchor=north west] at ($(causal.south west)+(0,-1mm)$) {association: identities change only};
  \end{tikzpicture}
  \caption{System overview.  The 3D detector is a frozen public checkpoint
  run at two input scales after gauge canonicalization; class-wise source
  selection is the only stage that changes boxes.  All later stages are
  asserted to change object identities only.  Blue stages are shared with
  GaugeAlign~\cite{wu2026gaugealign}; the $0.75\times$ pass, the source
  selection, and every orange stage are ours.}
  \label{fig:pipeline}
\end{figure}
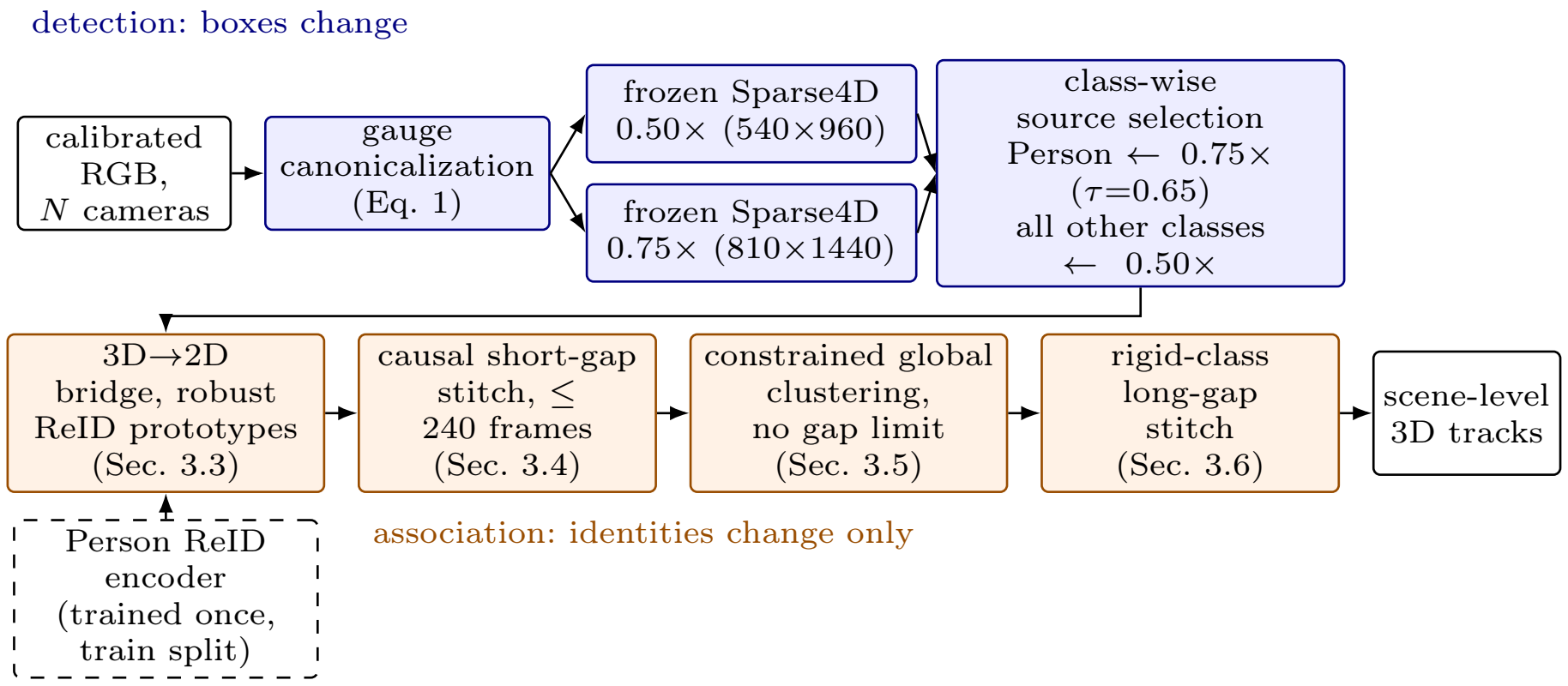

Figure~\ref{fig:pipeline} gives the pipeline.  Sections~\ref{sec:gauge} and
\ref{sec:multiscale} produce boxes; Sections~\ref{sec:bridge}--\ref{sec:rigid}
change only identities.

\subsection{Frozen Detector and Gauge Canonicalization}
\label{sec:gauge}

We use the public NVIDIA TAO Sparse4D warehouse checkpoint (release 2.2,
ResNet-50 backbone, 900 anchors) and never fine-tune it.  Let calibrated camera $i$ have extrinsics
$[R_i\mid t_i]$ and world-frame centre $C_i=-R_i^\top t_i$.  Following
GaugeAlign~\cite{wu2026gaugealign}, the network is centred before inference,
\begin{equation}
  \bar C=\frac{1}{N}\sum_{i=1}^{N} C_i, \qquad
  t_i'=t_i+R_i\bar C,
  \label{eq:gauge}
\end{equation}
and every predicted centre $p'$ is mapped back by $p=p'+\bar C$; dimensions,
yaw, class, and instance identity are untouched.  We also keep GaugeAlign's
class-taxonomy mapping, its temporal confidence decay of 0.95 for the
instance bank, and its floor-region filter.  Within each pass the detector's
instance identities are kept as raw tracklets.

\subsection{Class-Wise Multiscale Source Selection}
\label{sec:multiscale}

The checkpoint was trained on $540\times960$ inputs, half the native
$1920\times1080$ resolution, and the production pipeline runs it there.
Matching misses to ground truth on validation scene~20 shows where recall is
lost: 46\% of the missed Person frames are below 40~pixels tall in every
camera and another 45\% are 40--80~pixels; 54\% are visible in only one
camera.  Running the same checkpoint on $810\times1440$ inputs ($0.75\times$)
raises raw Person HOTA on a validation segment from 39.39 to 45.32 but
collapses Forklift (57.78 to 36.32) and PalletTruck (75.42 to 41.51), whose
large extents fall outside the anchor statistics the model learned
(Fig.~\ref{fig:scale}a).  A union of both passes creates duplicate boxes for
every object both passes see.

We therefore select the source per class.  For input scale $s$, scene $w$,
and class $c$ let
\begin{equation}
  \mathcal B_{w,c}^{s}(\tau_c)
  =\{b\in\mathcal B_w^s: y(b)=c,\ q(b)\geq\tau_c\},
  \label{eq:source}
\end{equation}
where $q$ is detector confidence.  The final box set for $(w,c)$ is exactly
one $\mathcal B_{w,c}^{s}(\tau_c)$: Person uses $s=0.75$ with
$\tau=0.65$ (chosen on validation, Sec.~\ref{sec:scaleexp}); Forklift,
NovaCarter, Transporter, FourierGR1T2, AgilityDigit, and PalletTruck use
$s=0.50$ with the production thresholds (0.7 for Forklift, 0.6 for
PalletTruck, 0.2 otherwise).  Because each class comes from one pass, raw
instance identities remain consistent within a class and no deduplication is
needed.  Source selection precedes every association stage, so appearance
descriptors are built for the selected rows.  A variant that also takes the
three humanoid and robot classes from $0.75\times$ was built but not scored on
the server (Sec.~\ref{sec:negative}).

\subsection{Geometry-to-Appearance Bridge}
\label{sec:bridge}

Sparse4D's instance identities do not survive long occlusions, so we attach
image appearance to its 3D tracks through the calibration.

\paragraph{ReID encoder.}
The encoder is an ImageNet-pretrained ResNet-50~\cite{he2016resnet} with a
512-dimensional linear--batch-norm head~\cite{luo2019bagoftricks} and $\ell_2$
normalization.  Training crops are cut from the 20 labelled training scenes
at a 15-frame stride: 290{,}592 Person crops of 621 identities, resized to
$256\times128$.  Batches hold 32 identities with four images each; the loss is
\begin{equation}
  \mathcal L=\mathcal L_{\mathrm{CE}}+2\,\mathcal L_{\mathrm{BH}}(m{=}0.4),
  \label{eq:reidloss}
\end{equation}
with $\mathcal L_{\mathrm{BH}}$ the batch-hard triplet
loss~\cite{hermans2017triplet}, AdamW (learning rate $3\times10^{-4}$,
weight decay $10^{-4}$), and random erasing~\cite{zhong2020randomerasing}
with probability 0.4.  An epoch is one identity-balanced pass over the 621
identities (20 batches of 128 crops); training runs 60 such epochs, 1{,}200
iterations in about 15 minutes on one H100.  The recipe was chosen among
four by nearest-neighbour rank-1 on a held-out 10{,}000-crop subset; the
final checkpoint reaches 90.03\% (peak 90.34\%).  Degraded training twins
(Sec.~\ref{sec:data}) are part of the crop set.

\paragraph{Projection and matching.}
For every Person box we project its eight 3D corners into every camera that
sees it and take the enclosing rectangle.  Within each
scene--frame--camera tuple, projected boxes are matched one-to-one by the
Hungarian algorithm~\cite{kuhn1955hungarian} to the cached 2D person
detections of the ReID extractor with cost
\begin{equation}
  c(p,d)=0.7\,\bigl(1-\operatorname{IoU}(p,d)\bigr)
  +0.3\,\frac{\min(\delta_{\mathrm{ctr}}(p,d),2)}{2},
  \label{eq:bridgecost}
\end{equation}
where $\delta_{\mathrm{ctr}}$ is the centre distance normalized by the
detection diagonal.  A pair is a candidate if IoU $\ge 0.02$ or
$\delta_{\mathrm{ctr}}\le1$, and is accepted when $c\le0.82$.  On the test
set this yields 5{,}450{,}393 camera projections and at least one accepted
match for 820{,}383 of the 832{,}033 Person rows (98.60\%).

\paragraph{Robust aggregation.}
For aggregation a match must additionally have detection confidence
$\ge0.15$, IoU $\ge0.10$, and cost $\le0.72$; its descriptor is weighted by
\begin{equation}
  \omega=\sqrt{q}\,(0.25+0.75\operatorname{IoU})\,\exp(-\delta_{\mathrm{ctr}}).
  \label{eq:weight}
\end{equation}
Simultaneous camera observations are first fused into one descriptor per 3D
row.  Per raw tracklet, a provisional weighted mean is formed, the 10\% of
rows least similar to it are dropped (never below a similarity of 0.20), and
the mean is recomputed to give a global prototype; start and end prototypes
use the first and last 90 frames.  The test submission has 977 tracklet
prototypes.

\subsection{Short-Gap Causal Stitching}
\label{sec:short}

For each tracklet we fit a ground-plane velocity to its last 15 observations
and clip its magnitude to 2~m/s.  A predecessor $a$ that ended $\Delta f$
frames before successor $b$ started is a candidate if
\begin{equation}
  \bigl\|x_b^{\mathrm{start}}
  -\bigl(x_a^{\mathrm{end}}+v_a\Delta f\bigr)\bigr\|_2
  \leq 0.35+2\,\frac{\Delta f}{30}\ \text{m},
  \label{eq:motiongate}
\end{equation}
with $\Delta f\le240$, an absolute residual of at most 2~m, a mean
log-size change of at most 0.35, and, when both prototypes exist, an
end-to-start ReID cosine similarity of at least 0.625.  Motion, gap, size,
and appearance terms are normalized and summed, and the assignment at each
successor start frame is solved by the Hungarian algorithm.  Only tracklets
that have already ended are considered, so this stage is causal.  It makes
133 joins in the test submission and is retained because it is cheap and
raises the quality of the prototypes that feed the next stage.

\subsection{Constraint-Aware Global Track Consolidation}
\label{sec:global}

\begin{figure}[t]
  \centering
  \includegraphics[width=0.8\textwidth]{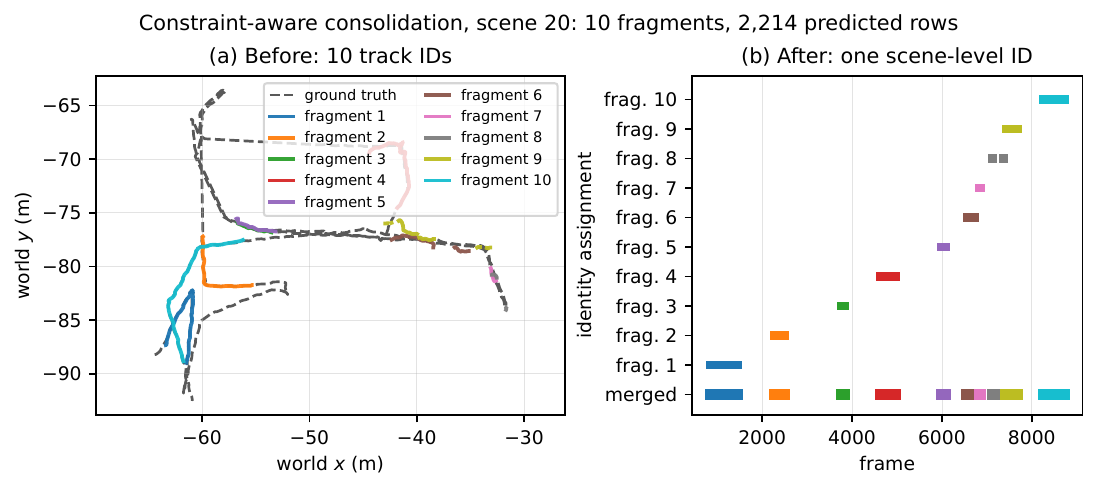}
  \caption{A fixed-box validation example (scene~20).  Before global
  clustering, ten predicted identities cover one ground-truth person (dashed);
  the constrained global stage merges them into one scene-level identity.  The
  right panel shows that only labels change: the temporal support of every
  fragment is preserved, and no two fragments overlap in time.}
  \label{fig:association-example}
\end{figure}

Figure~\ref{fig:gaps} shows that most same-identity gaps are longer than any
reasonable causal horizon.  We therefore consolidate the whole Person tracks
of a scene offline.  Each track $T$ is a node with the row-weighted centroid
of its raw tracklets' global prototypes,
\begin{equation}
  e_T=\frac{\sum_{k\in T}n_k e_k}
  {\bigl\|\sum_{k\in T}n_k e_k\bigr\|_2},
  \label{eq:centroid}
\end{equation}
where $n_k$ counts the rows tracklet $k$ contributes; tracks with fewer
than 30 rows are not eligible.  We repeatedly take the live pair with the
highest cosine similarity and merge it if the similarity is at least
$\tau=0.65$ and both hard constraints hold:
\begin{enumerate}
  \item \textbf{Temporal exclusion}: the two tracks share no frame.
  \item \textbf{3D reachability}: in the merged, time-ordered sequence, every
  switch from one track to the other spanning $\Delta f$ frames covers a
  ground-plane distance of at most $2.0+2.0\,\Delta f/30$ metres.
\end{enumerate}
After a merge the centroid is updated and the search continues; a pair that
fails a constraint is excluded for good.  There is no gap limit, so the
procedure uses future observations and is offline.  It makes 387 merges in
the test submission.  The constraints matter because appearance is weakest
precisely where it is most needed: warehouse workers wear similar clothing,
and in the restyled scenes distinct people routinely exceed $\tau$.  Two
such people who are visible at the same time are certainly different, and
temporal exclusion turns that into a free, exact veto; on validation it
refuses more than 600 above-threshold pairs.  The reachability gate bounds
the remaining risk of joining people who are never seen together.
Section~\ref{sec:constraints} measures both; Fig.~\ref{fig:association-example}
shows one consolidated identity.

\subsection{Class-Aware Long-Gap Stitching for Rigid Classes}
\label{sec:rigid}

For the six non-Person classes, a Person-trained appearance model is not
informative, but these objects are rigid, their boxes are consistent, and
their motion is slow.  We reuse the causal stitcher of Sec.~\ref{sec:short}
with a 1{,}800-frame gap and class-specific gates in place of
Eq.~\eqref{eq:motiongate}: Forklift and Transporter use a 3~m base gate and
3~m/s; PalletTruck, NovaCarter, FourierGR1T2, and AgilityDigit use 1~m and
2~m/s, and PalletTruck additionally caps the absolute residual at 4~m.  The
gates were selected on validation scene~22 and on training scenes 10 and 15,
which contain the rare classes (Sec.~\ref{sec:rare}).  In the test submission
these stages make 152 joins (53 Forklift, 4 PalletTruck, 1 NovaCarter,
10 Transporter, 26 FourierGR1T2, 58 AgilityDigit).  Every association stage
asserts numerically that frames, classes, positions, and dimensions are
identical to its input and that no identity occupies a frame twice.

\section{Experiments}
\label{sec:experiments}

\subsection{Data, Metric, and Protocol}
\label{sec:data}

\begin{table}[t]
  \centering
  \caption{Scenes used in this paper.  Ground-truth identity counts are given
  for the validation classes that have annotations; FourierGR1T2 and
  AgilityDigit occur only in the test split (and in training scenes).
  Validation scene~021 is a generatively restyled rendering of scene~020 with
  identical ground truth; test scenes 023/024 form the same kind of pair.}
  \label{tab:data}
  \scriptsize
  \resizebox{\textwidth}{!}{%
  \begin{tabular}{llrrrrrrr}
    \toprule
    Split & Scene & Cameras & Frames & Person & Forklift & PalletTruck & NovaCarter & Transporter \\
    \midrule
    train & 000--019 & 4--19 & 9{,}000 & \multicolumn{5}{l}{ReID training (621 identities); gates (010, 015)} \\
    \midrule
    val & 020 & 16 & 9{,}000 & 60 & 9 & 10 & -- & -- \\
    val & 021 (degraded twin) & 16 & 9{,}000 & 60 & 9 & 10 & -- & -- \\
    val & 022 & 4 & 9{,}270 & 24 & 1 & -- & 2 & 2 \\
    \midrule
    test & 023 / 024 (twins) & 20 / 20 & 9{,}000 & \multicolumn{5}{l}{hidden} \\
    test & 025 & 10 & 9{,}000 & \multicolumn{5}{l}{hidden; all rare classes present} \\
    test & 026 / 027 & 4 / 7 & 1{,}800 & \multicolumn{5}{l}{hidden} \\
    \bottomrule
  \end{tabular}}
\end{table}

Table~\ref{tab:data} summarizes the data~\cite{wang2026physicalai}.  Each
scene provides synchronized $1920\times1080$ video at 30~fps, calibration, and
a map; training and validation scenes also provide 2D/3D annotations.  Depth
is not used.  The seven classes are Person, Forklift, NovaCarter, Transporter,
FourierGR1T2, AgilityDigit, and PalletTruck.  Validation has 144 Person
identities but only 19 Forklift, 20 PalletTruck, two NovaCarter, and two
Transporter identities, and no humanoid robots, so rare-class decisions were
cross-checked on training scenes 10 and 15 with the production detector.

Hidden-test scores come from the official evaluation server.  Local scores
use the challenge's TrackEval adaptation~\cite{luiten2020trackeval} with its
3D IoU matching; to iterate on identity-only changes we cache the per-frame
3D IoU matrices once and re-run TrackEval's HOTA aggregation for every
relabelling.  This cached evaluator reproduces the official script digit for
digit on every scene and class we checked.  Local figures are combined
over classes by detection count unless noted; the server's class and scene
weighting is not published, so validation is used for controlled comparisons
and the server only for end-to-end submissions.

\subsection{Implementation and Cost}

Both detector passes use the same checkpoint, anchors, confidence decay, and
floor filter; the $0.75\times$ pass changes only the input dimension.  On one
H100 with six decoding workers, 1{,}800 frames of 16 cameras take 442, 467,
565, and 647~s at $0.625\times$, $0.75\times$, $0.875\times$, and
$1.0\times$.  All post-detection stages for the whole test set (1.26 million
rows) run in about 20 minutes on 16 CPU cores, of which global clustering
takes well under a minute; the association scripts are pure NumPy.
Association hyperparameters were selected on validation and held fixed on
test.

\subsection{Evaluation-Server Results}

\begin{table}[t]
  \centering
  \caption{Hidden-test results from the official evaluation server (percent).
  Top: the official Track~1 online ranking at the close of the challenge, as
  reported by the respective teams~\cite{wu2026gaugealign,tran2026syn2realtrack,shrestha2026sparseqid}.
  Bottom: the general leaderboard snapshot of 30~October~2026 after our final
  submission, and the updated ELMO entries of 3 and 6~November~2026.  Our
  entry is offline and post-deadline; it is a server rank, not a claim to
  the challenge award or the online bonus.}
  \label{tab:leaderboard}
  \small
  \resizebox{\textwidth}{!}{%
  \begin{tabular}{llrrrr}
    \toprule
    Ranking & System & HOTA & DetA & AssA & LocA \\
    \midrule
    official (online) & GaugeAlign (EVA)~\cite{wu2026gaugealign} & 56.5447 & 55.6444 & 49.3929 & 79.5558 \\
    & Syn2RealTrack (SKKU-AL-T1)~\cite{tran2026syn2realtrack} & 52.0118 & 45.3056 & 56.5047 & 76.2410 \\
    & SparseQID (Playbox)~\cite{shrestha2026sparseqid} & 38.0105 & 40.2592 & 31.0978 & 75.1778 \\
    \midrule
    general leaderboard & \textbf{ParkLabs (ours)} & 62.3947 & 57.4501 & 56.8131 & 80.1272 \\
    (30 Oct 2026) & ELMO & 61.8129 & 60.1596 & 54.1241 & 80.8056 \\
    & Metropolis & 60.7019 & -- & -- & -- \\
    \midrule
    (3 Nov 2026) & ELMO, updated entry & 62.6375 & 60.1570 & 55.3186 & 80.8063 \\
    (6 Nov 2026) & ELMO, updated entry & 65.8422 & 63.8761 & 57.1033 & 81.9793 \\
    \bottomrule
  \end{tabular}}
\end{table}

\begin{table}[t]
  \centering
  \caption{Our evaluation-server progression.  Rows are submitted systems in
  order, not independent single-factor ablations.  Rows~2--6 change
  identities only (DetA constant to the third decimal); row~7 changes the
  Person box source as well.}
  \label{tab:progression}
  \small
  \setlength{\tabcolsep}{4pt}
  \resizebox{\textwidth}{!}{%
  \begin{tabular}{llrrrrr}
    \toprule
    & Configuration & HOTA & DetA & AssA & LocA & $\Delta$HOTA \\
    \midrule
    1 & GaugeAlign reproduction (frozen Sparse4D, $0.50\times$) & 56.7690 & 55.7189 & 49.6755 & 79.5895 & -- \\
    2 & + ReID gate in the causal stitcher & 56.8737 & 55.7188 & 49.8028 & 79.5895 & +0.1047 \\
    3 & + class-wise score thresholds, motion gate & 57.3536 & 56.5312 & 50.0756 & 79.8645 & +0.4799 \\
    4 & + causal ReID stitch, 180-frame gap & 57.6727 & 56.5311 & 50.4811 & 79.8646 & +0.3191 \\
    5 & + 240-frame gap (previous champion) & 57.7383 & 56.5314 & 50.5654 & 79.8645 & +0.0656 \\
    6 & + constrained global clustering, rigid-class stitching & 61.4704 & 56.5314 & 56.0554 & 79.8643 & +3.7321 \\
    7 & + Person from $0.75\times$, $\tau{=}0.65$ (\textbf{final}) & \textbf{62.3947} & \textbf{57.4501} & \textbf{56.8131} & \textbf{80.1272} & +0.9243 \\
    \bottomrule
  \end{tabular}}
\end{table}

Table~\ref{tab:leaderboard} places the result.  Our reproduction of the
official winner (56.7690) is slightly above the authors' own 56.5447, which
we attribute to threshold details of the release.  On the 30~October
snapshot our system led ELMO by 0.5818 HOTA.  ELMO's subsequent entries of
3 and 6~November reached 62.6375 and 65.8422; the second step added 3.7 DetA
and 1.2 LocA points at once, which indicates a detector change.  In every
snapshot ELMO holds the detection advantage (2.7, 2.7, and 6.4 DetA points)
while ours was association (2.69 and 1.49 AssA points, now $-0.29$).  This
is the paper's premise in one table: once the detector is strong,
association decides the order, and once both detectors are strong,
detection decides it again.  The frozen-detector route of this paper has
reached the point where the detector itself is the limit.

Table~\ref{tab:progression} separates the effects.  Rows 2--5 are the
causal-stitcher refinements of the production system; together they are
worth 0.97 HOTA, and the last of them only 0.07, which is the diminishing
return that motivated an offline stage.  Row~6, constrained global
clustering plus rigid-class stitching, adds 3.7321 HOTA and 5.49 AssA while
DetA is unchanged to four decimals, so the whole gain is identity topology.
Row~7, the $0.75\times$ Person source, adds 0.9187 DetA, 0.7577 AssA, and
0.2629 LocA; its association gain is indirect, since more complete Person
observations yield longer tracklets and better prototypes.  Validation
showed a larger detection-weighted gain for the same change (+3.2
det-averaged HOTA over the five annotated classes), which is consistent with
a server weighting in which Person is one class among seven and with test
scenes 25--27 containing few people.

\subsection{Controlled Validation: Stages, Fixed Boxes}
\label{sec:stages}

\begin{table}[t]
  \centering
  \caption{Stage-wise validation results on the final box set (Person from
  $0.75\times$ at $\tau{=}0.65$, other classes from $0.50\times$).  Within
  each class only object identities change between columns, so DetA and LocA
  are constant.  ``IDs'' is the number of predicted identities; the number of
  ground-truth identities is in parentheses.  Person passes through the
  causal and the global stage; rigid classes pass through the long-gap
  stitch.}
  \label{tab:stages}
  \small
  \setlength{\tabcolsep}{3.5pt}
  \begin{tabular}{lrrrrrrrrr}
    \toprule
    & & \multicolumn{3}{c}{raw Sparse4D IDs} & \multicolumn{2}{c}{+ causal stitch} & \multicolumn{3}{c}{+ global / long-gap} \\
    \cmidrule(lr){3-5}\cmidrule(lr){6-7}\cmidrule(lr){8-10}
    Class & GT IDs & HOTA & AssA & IDs & HOTA & IDs & HOTA & AssA & IDs \\
    \midrule
    Person & 144 & 29.189 & 20.887 & 1{,}210 & 31.009 & 1{,}041 & \textbf{40.108} & 39.219 & 516 \\
    Forklift & 19 & 46.627 & 50.553 & 124 & -- & -- & \textbf{49.655} & 56.901 & 25 \\
    PalletTruck & 20 & 58.398 & 67.709 & 33 & -- & -- & \textbf{60.936} & 73.752 & 22 \\
    NovaCarter & 2 & 16.978 & 8.381 & 10 & -- & -- & \textbf{28.038} & 22.814 & 3 \\
    Transporter & 2 & 16.979 & 11.472 & 14 & -- & -- & \textbf{25.251} & 23.943 & 4 \\
    \bottomrule
  \end{tabular}
\end{table}

Table~\ref{tab:stages} isolates association on the final boxes.  The causal
stitch recovers 1.8 Person HOTA; constrained global clustering recovers a
further 9.1, reducing 1{,}041 identities to 516 for 144 people, and raising
AssA from 23.5 to 39.2.  Per scene, Person HOTA reaches 46.24 on scene~20,
31.58 on its degraded twin, and 44.58 on scene~22.  The remaining excess of
predicted over true identities is deliberate: with the constraints in place,
lowering $\tau$ further merges fewer true pairs than it damages
(Sec.~\ref{sec:constraints}).  The same long-gap principle gives rigid
classes +3.0 (Forklift), +2.5 (PalletTruck), +11.1 (NovaCarter), and +8.3
(Transporter) HOTA from motion and box consistency alone.  On the
half-resolution boxes of the previous champion the same Person stage moved
HOTA from 26.849 to 35.115 (1{,}046 to 511 identities), which is the
validation counterpart of row~6 in Table~\ref{tab:progression}.

\subsection{Input Scale and Threshold}
\label{sec:scaleexp}

\begin{figure}[t]
  \centering
  \includegraphics[width=\textwidth]{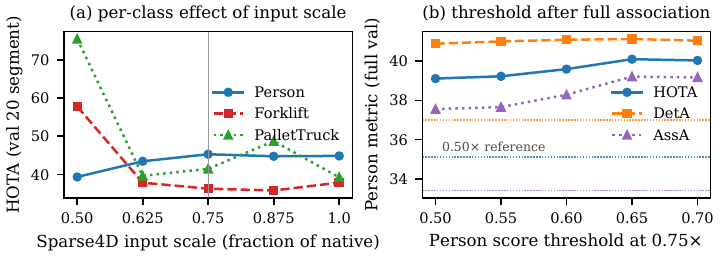}
  \caption{(a)~Raw per-class HOTA on validation scene~20, frames 0--1{,}799,
  as a function of the Sparse4D input scale (exact evaluation; only the input
  dimension changes).  Person improves up to $0.75\times$ and then saturates;
  both large classes collapse away from the $0.50\times$ training scale.
  (b)~Full-validation Person HOTA, DetA, and AssA after the complete
  association chain for the $0.75\times$ source at five score thresholds;
  dotted lines are the $0.50\times$ source with the same chain.}
  \label{fig:scale}
\end{figure}

Figure~\ref{fig:scale}a sweeps five input scales on a validation segment.
Person HOTA rises from 39.39 at the training scale to 43.48, 45.32, 44.79,
and 44.89 at $0.625\times$, $0.75\times$, $0.875\times$, and native
resolution, while Forklift and PalletTruck lose 20--35 points at every scale
other than $0.50\times$.  Native resolution is thus not automatically best:
the middle scale balances small-object evidence against the image statistics
the checkpoint learned, and it costs 28\% less than $1.0\times$.
Figure~\ref{fig:scale}b then fixes the source and varies the Person threshold
through the whole association chain on the full validation set: HOTA is
39.118, 39.234, 39.601, 40.108, and 40.044 at thresholds 0.50 to 0.70 in
steps of 0.05, against 35.115 for the half-resolution source, while LocA
rises monotonically from 78.41 to 78.94.  The optimum is broad: every
threshold beats the half-resolution source by at least 4.0 HOTA, and 0.65
is best by 0.06 over 0.70, with the same ordering on each of the three
scenes.  The threshold interacts with association: a higher threshold raises
raw precision but removes the low-confidence rows that bridge fragments,
which is why the raw optimum on the segment (0.70) is not the chain optimum.

\subsection{Constraint Ablation and Sensitivity}
\label{sec:constraints}

\begin{table}[t]
  \centering
  \caption{Constraint ablation for global Person clustering on the final
  validation boxes (identities only change; exact evaluation).  Every variant
  keeps temporal exclusion, without which the output would assign one
  identity to two boxes of the same frame.  ``Merges'' and ``refused''
  (candidate pairs above $\tau$ that a constraint rejected: temporal
  exclusion / reachability) are summed over the three scenes.}
  \label{tab:constraints}
  \small
  \setlength{\tabcolsep}{3.5pt}
  \resizebox{\textwidth}{!}{%
  \begin{tabular}{llrrrrrrr}
    \toprule
    Variant & $\tau$ & HOTA & AssA & AssRe & AssPr & IDs & merges & refused \\
    \midrule
    none (after causal stitch) & -- & 31.009 & 23.523 & 25.086 & 72.281 & 1{,}041 & 0 & -- \\
    \midrule
    temporal exclusion only & 0.55 & 40.084 & 39.169 & 44.641 & 60.871 & 414 & 627 & 1{,}077 / -- \\
    temporal exclusion only & 0.65 & \textbf{40.134} & \textbf{39.269} & 44.316 & 63.103 & 498 & 543 & 618 / -- \\
    temporal exclusion only & 0.75 & 39.421 & 37.905 & 42.292 & 66.238 & 628 & 413 & 234 / -- \\
    temporal exclusion only & 0.85 & 37.559 & 34.455 & 38.059 & 68.672 & 786 & 255 & 86 / -- \\
    \midrule
    + reachability, tight (0.5~m, 1~m/s) & 0.65 & 39.292 & 37.663 & 41.815 & 65.586 & 565 & 476 & 617 / 182 \\
    + reachability, \textbf{production} (2~m, 2~m/s) & 0.65 & 40.108 & 39.219 & 44.047 & 63.849 & 516 & 525 & 606 / 52 \\
    + reachability, loose (5~m, 5~m/s) & 0.65 & 40.128 & 39.257 & 44.165 & 63.472 & 505 & 536 & 635 / 26 \\
    \midrule
    production, every track eligible & 0.65 & 40.097 & 39.199 & 44.023 & 63.834 & 492 & 549 & 621 / 55 \\
    production, average linkage & 0.65 & 39.606 & 38.258 & 42.676 & 64.519 & 558 & 483 & 291 / 21 \\
    \bottomrule
  \end{tabular}}
\end{table}

\begin{figure}[t]
  \centering
  \includegraphics[width=\textwidth]{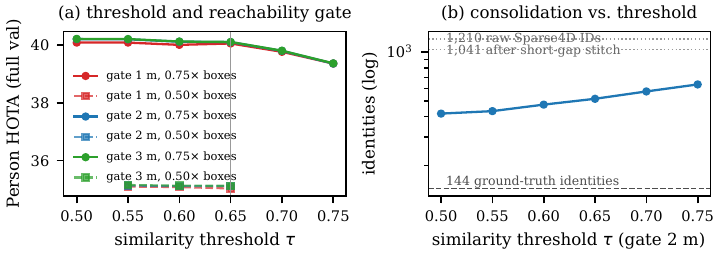}
  \caption{Sensitivity of constrained global clustering.  (a)~Full-validation
  Person HOTA against the similarity threshold $\tau$ for three reachability
  base gates, on the final $0.75\times$ boxes (circles) and on the previous
  half-resolution boxes (squares).  (b)~Predicted Person identities against
  $\tau$ at the production gate; the dotted lines mark the input identity
  counts and the dashed line the 144 ground-truth identities.}
  \label{fig:sensitivity}
\end{figure}

Table~\ref{tab:constraints} varies the constraints on the final boxes.
Temporal exclusion is the decisive one.  At the production setting it
refuses 606 candidate pairs whose similarity exceeds $\tau$, twelve times
as many as reachability refuses (52), and 459 of them are in the degraded
scene~21, where restyled textures make distinct people look alike.  Such
pairs are present in the same frame, so merging them is never right; the
counts measure how often appearance alone would have done so.  The
reachability gate, by contrast, is a safety margin rather than a
source of gain on this validation set: at $\tau=0.65$ removing it changes
HOTA by $+0.03$, loosening it to 5~m and 5~m/s by $+0.02$, and tightening it
to 0.5~m and 1~m/s costs 0.82, because genuine hand-offs across long
occlusions are refused (182 rejections).  Without the gate, $\tau$ can even
be lowered to 0.55 with almost no loss (40.084), at the price of a visibly
lower AssPr (60.9) and fewer identities than ground truth would warrant in
the degraded scene.  We keep the production gate because it bounds the
damage of a wrong merge at no measurable cost, and because the test scenes
(20 cameras) are larger than any validation scene.  Prototype eligibility is
immaterial (40.097 with every track eligible), and average linkage loses
0.50: its mean-pairwise criterion is more conservative, leaving 558
identities after 483 merges.

Figure~\ref{fig:sensitivity} shows the operating region.  Across the 36
settings on the final boxes with $\tau$ from 0.50 to 0.75, gates of 1, 2,
and 3~m, and a minimum of 15 or 30 rows, Person HOTA stays within
39.36--40.22; on the half-resolution boxes, 24 settings of $\tau$, gate, and
speed stay within 34.89--35.15.  The production setting ($\tau=0.65$, 2~m,
2~m/s, 30 rows) was chosen on the half-resolution boxes and transferred to
the $0.75\times$ boxes without retuning; retuning there would have gained
0.11.  The plateau towards lower $\tau$ is what temporal exclusion and the
trimmed prototypes buy: the harmful merges that a lower threshold admits are
overwhelmingly between people who are visible at the same time, and the
constraint removes them for free.

\subsection{Rare Classes}
\label{sec:rare}

\begin{table}[t]
  \centering
  \caption{Long-gap stitching for rigid classes.  Validation columns use all
  validation scenes (NovaCarter and Transporter occur only in scene~22).
  Training scenes 10 and 15 were run with the production detector, which was
  trained on the training split, so their absolute values are optimistic;
  the gates were fixed from this evidence and applied unchanged to the test
  submission.  Values are before $\to$ after stitching; ground-truth identity
  counts in parentheses.}
  \label{tab:rare}
  \small
  \resizebox{\textwidth}{!}{%
  \begin{tabular}{llllr}
    \toprule
    Class & Gate (base, speed) & Validation: HOTA & Validation: IDs (GT) & Train 010+015: HOTA \\
    \midrule
    NovaCarter & 1~m, 2~m/s & 16.978 $\to$ 28.038 & 10 $\to$ 3 (2) & 54.059 $\to$ 68.506 \\
    Transporter & 3~m, 3~m/s & 16.979 $\to$ 25.251 & 14 $\to$ 4 (2) & 48.210 $\to$ 55.859 \\
    FourierGR1T2 & 1~m, 2~m/s & no annotations & -- & 52.657 $\to$ 58.812 \\
    AgilityDigit & 1~m, 2~m/s & no annotations & -- & 60.829 $\to$ 63.115 \\
    Forklift & 3~m, 3~m/s & 46.627 $\to$ 49.655 & 124 $\to$ 25 (19) & -- \\
    PalletTruck & 1~m, 2~m/s, 4~m cap & 58.398 $\to$ 60.936 & 33 $\to$ 22 (20) & -- \\
    \bottomrule
  \end{tabular}}
\end{table}

If the server averages HOTA over classes, the four rare classes that appear
only in test scenes 25--27 weigh as much as Person, and they are heavily
fragmented there (our raw detector output has 74 AgilityDigit identities in
scene~25 for at most seven simultaneous objects).  Table~\ref{tab:rare}
reports the evidence behind the class-aware gates: every class gains 2--14
HOTA from motion-only stitching on held-out and training scenes, the
Transporter gate was widened to 3~m because scene~10 showed it beneficial
and scene~22 was indifferent, and a per-instance class majority vote across
the humanoid classes was tested and found neutral.  On the test submission
the rare-class stage reduces, for example, scene-25 FourierGR1T2 identities
from 40 to 14 and AgilityDigit from 74 to 17.

\subsection{Negative Results}
\label{sec:negative}

\begin{table}[t]
  \centering
  \caption{Alternatives that did not improve exact validation HOTA (Person
  unless stated).  ``Segment'' is validation scene~20, frames 0--1{,}799;
  ``raw'' means detector output before association.}
  \label{tab:negative}
  \scriptsize
  \setlength{\tabcolsep}{4pt}
  \begin{tabular}{>{\raggedright\arraybackslash}p{0.52\textwidth}>{\raggedright\arraybackslash}p{0.43\textwidth}}
    \toprule
    Alternative & Outcome \\
    \midrule
    \multicolumn{2}{l}{Association} \\
    Average linkage instead of centroid linkage & 34.41--34.58 vs 34.79--35.15 across the same thresholds \\
    Per-scene adaptive $\tau$ from ``certain-negative'' pairs (simultaneous tracks $\ge2$~m apart) & 31.45--33.88 vs 34.80; scene~22 holds look-alike people \\
    End-to-start ReID veto at every internal hand-off (0.2--0.6) & monotone loss, 35.11 to 32.77; endpoint crops are the most occluded \\
    Retuning $\tau$, gate, and minimum rows on the $0.75\times$ boxes & best 40.222 vs 40.108 transferred \\
    Linear gap interpolation inside merged tracks & only 8--11\% of missed frames lie in gaps $\le120$ frames \\
    Track-median box size or ground-plane smoothing & neutral or worse; Person length varies with pose \\
    Causal-stitcher gap beyond 240 frames, reverse-motion gates & no gain; reverse gates lose 1--2 HOTA (server rows 4--5) \\
    \midrule
    \multicolumn{2}{l}{Detection} \\
    Union of $0.50\times$ and $0.75\times$ outputs (scaled rows added where no $0.50\times$ row lies within 0.5~m) & Person 38.89 vs 39.39 ($0.50\times$) and 45.32 ($0.75\times$); Forklift 51.11 vs 57.78; PalletTruck 64.52 vs 75.42 (segment) \\
    Track-level fusion of $0.50\times$ Person tracks not covered by $0.75\times$ & 45.02--45.28 vs 45.32 (segment), 9 variants \\
    Input scales $0.625\times$, $0.875\times$, $1.0\times$ for Person & 43.48, 44.79, 44.89 vs 45.32 (segment) \\
    Instance-bank confidence decay 0.90 / 0.98 at $0.75\times$ & 45.39 / 44.71 vs 45.39 (segment, $\tau{=}0.65$) \\
    Lower Forklift / PalletTruck thresholds (0.6) & +0.14 / 0.00 HOTA; production kept \\
    Transporter, FourierGR1T2, AgilityDigit from $0.75\times$ (variant D-prime) & val Transporter +0.95 but the $0.75\times$ pass loses all robot detections in test scene~26; built, not scored \\
    Per-instance class majority vote among Person and humanoids & $\pm0.2$ HOTA \\
    2D confirmation of low-score 3D candidates (0.05--0.6) & $\approx50\%$ false positives after confirmation; covers 14\% of misses \\
    Broad recovery of low-confidence 3D rows & lower HOTA (production system) \\
    ResNet-101 Sparse4D variant & 41.43 HOTA on the server \\
    Fine-tuning the checkpoint on five training scenes at $0.75\times$ (LR 5e-6 / 2e-5, with and without the background-query loss, with and without a frozen backbone; 1{,}000 iterations) & Person 40.0--43.8 vs 45.4 at every checkpoint (segment, raw); LocA $-1$ to $-3.5$; only Forklift / PalletTruck adapt to $0.75\times$ \\
    Fine-tuning on validation scenes 20--21, tested on scene 22 & Person 44.6--45.3 vs 46.9 on scene 22; 43.4--44.4 vs 45.4 in-sample on scene 20 \\
    \bottomrule
  \end{tabular}
\end{table}

Table~\ref{tab:negative} lists what was tried and rejected, with the exact
validation outcome.  Three lessons shaped the final design.  Refinements
that try to extract more from appearance, such as adaptive thresholds,
average linkage, and junction vetoes, all lose, whereas the one constraint
that matters is geometric, exact, and free.  Detection cannot be improved by
adding rows: every form of union, fusion, or low-confidence recovery lost to
replacing the source for one class, and the union of both scales is worse
than either alone.  And the detector itself resists improvement by
fine-tuning: the last two rows of Table~\ref{tab:negative} train the
released checkpoint with the TAO training stack at a batch-matched learning
rate (the released rate collapses the model, as GaugeAlign
reports~\cite{wu2026gaugealign}), and Person is lower at every checkpoint
of every recipe, on held-out scenes and even on the frames it was trained
on, while the large classes adapt readily to the $0.75\times$ scale.  The
released checkpoint scores 69 Person HOTA on a training scene against
39--45 on validation, so the training split is in its training data and
re-training on it adds no information; the small, distant people that
account for most misses are not recovered by this model at this capacity.

\section{Limitations and Reproducibility}
\label{sec:limitations}

Global clustering uses future frames, so the system is offline and
ineligible for the online bonus; a causal variant would need to forgo the
long-gap merges that account for most of the gain.  The leaderboard result
is a hidden-test measurement with no per-class or per-scene breakdown, and
local validation has three sequences, two of which are the same world.
Hyperparameters are fixed class-specific gates; they sit on broad plateaus
(Fig.~\ref{fig:sensitivity}) but a learned merge probability with the same
hard constraints would transfer more gracefully.  The largest remaining
headroom is detection: the leading entry's DetA is now 6.4 points higher
than ours, and our own Person recall on validation is 45.5\% at 69.1\%
precision.  Centroid
linkage can propagate a wrong merge; the constraints bound but do not
eliminate this, and we observed AssPr fall as AssRe rises
(Table~\ref{tab:stages}).  Finally, the paired clean and restyled scenes
(Table~\ref{tab:data}) share a world; we treat every test scene
independently, and did not attempt cross-scene fusion.

\paragraph{Reproducibility.}
\label{sec:repro}
\begin{center}\small
  scored submission: \texttt{agent2\_globalreid\_D\_fd075p065\_C2/track1.zip}\\
  SHA-256 \texttt{92581abca47eb8108c60fec52452885d441e10bff2bc43acc36aac4bf3b2e7d4}
\end{center}
\noindent The archive contains one file, \texttt{track1.txt}, with 1{,}264{,}782 rows
(832{,}033 Person, 143{,}932 Forklift, 20{,}132 NovaCarter, 46{,}827
Transporter, 24{,}890 FourierGR1T2, 41{,}520 AgilityDigit, 155{,}448
PalletTruck), no duplicate (scene, class, identity, frame) key, and no row
dropped by the strict packager.  Its non-Person rows are identical as a
multiset to the previous champion's, and its non-Person identity partition
is identical to that of row~6 in Table~\ref{tab:progression}; these
properties were asserted at build time.  Every stage is deterministic given
its inputs; we keep the input hashes, configuration scripts, merge logs,
per-stage outputs, and the cached-evaluator JSON files for every number in
this paper.  The detector is the public NVIDIA checkpoint; the ReID encoder
is trained locally in 15 minutes from the released annotations.  Public
repository details and a license table will accompany the final release.

\section{Conclusion}

We presented a detector-preserving system for calibrated multi-camera 3D
tracking that reaches \final{} \hota{} on the AI City Challenge 2026 Track~1
evaluation server, first on the general leaderboard snapshot of 30~October
2026; a competitor's later detector update has since moved the lead to
65.84, which marks the limit of the frozen-detector route: fine-tuning the
released checkpoint did not raise Person detection under any recipe we
tried, so the next step is a different detector, not a better use of this
one.  The frozen
Sparse4D detector was not replaced or fine-tuned; the gain came from choosing
the input scale per class and from repairing the topology of the detector's
tracks with appearance under hard physical constraints.  Two results seem
general.  Replacing a class's source at a second scale is safer than any
form of multi-scale union for a query-based detector with learned size
priors.  And offline ReID clustering of 3D tracks becomes reliable once
simultaneous presence is enforced as a hard veto, because it supplies
exactly the negative evidence that appearance lacks among similarly dressed
people; a metric reachability gate then bounds the remaining risk at no
cost.  For calibrated 3D tracking, repairing a strong detector's
trajectories can be as consequential as replacing the detector.

\bibliographystyle{splncs04}
\bibliography{parklabs_track1}

\end{document}